\documentclass[sigconf]{acmart}
\usepackage{booktabs}
\usepackage{graphicx}

\acmConference[AI in Health '26]{UT Austin AI in Health}{Spring 2026}{Austin, TX}

\begin{document}

\title{Faithful by Design: Evaluating and Improving LLM-Generated Clinical Trial Summaries for Multi-Stakeholder Audiences}

\author{Robert Williams}
\affiliation{%
  \institution{University of Texas at Austin}
  \department{Computer \& Data Science Online}
  \city{Austin}
  \state{Texas}
  \country{USA}
  }

\email{rgw@utexas.edu}

\begin{abstract}
Large language models are increasingly used to summarize clinical trial results for healthcare providers, patients, and payers, but their tendency to hallucinate poses significant risks in this high-stakes context. This study introduces a benchmark evaluation framework for measuring the faithfulness of LLM-generated clinical trial summaries across three stakeholder audiences. The framework consists of 200 stratified trials drawn from the Aggregate Analysis of ClinicalTrials.gov database, evaluated using audience-specific prompt templates and a six-dimension faithfulness annotation schema. Baseline measurements were established for GPT-4o, Claude Sonnet 4.6, and Gemini 2.5 Flash across 1,800 generated summaries scored using a cross-encoder natural language inference (NLI) model. Unsupported Claims was identified as the dominant failure mode across all three models, with a mean annotation score of 1.55 out of three. A knowledge-graph-augmented retrieval system grounded in the PubMed Knowledge Graph was developed and evaluated against the baseline, producing statistically significant improvements in NLI-based faithfulness scores (entailment $+0.0125$, faithfulness $+0.0130$, $p < 0.0001$). Improvement pathways were model-dependent, with GPT-4o improving primarily through contradiction reduction while Claude Sonnet 4.6 and Gemini 2.5 Flash improved through increased entailment.
\end{abstract}



\maketitle

\section{Introduction}

Clinical trial results are among the most consequential documents in healthcare. They underpin regulatory approvals, inform prescribing and treatment decisions, guide coverage determinations by payers, and must be communicated to patients in plain language to support informed consent. A single completed trial may require distinct summaries for healthcare providers, patients, and payers, each demanding a different framing of the same underlying evidence. ClinicalTrials.gov currently lists more than 400,000 registered trials, and this summarization workload currently falls to skilled medical writers who must translate complex findings for each audience without distorting them.

Large language models (LLMs) are natural candidates for automating this process. They can synthesize complex scientific text, adapt tone and framing for different audiences, and operate at a speed and scale that human writers cannot match~\cite{landman2024llm}. However, LLMs are known to hallucinate, producing outputs that are unfaithful to their source material despite appearing fluent and coherent~\cite{min2023factscore}. In a clinical summarization study of 12,999 clinician-annotated sentences, Asgari et al.\ found a 1.47\% hallucination rate, with 44\% of those hallucinations classified as major, meaning they carried real potential to affect patient diagnosis or clinical management~\cite{asgari2025framework}. In the context of clinical trial communication, even a low hallucination rate is unacceptable: a fabricated effect size, a suppressed adverse event, or an overstated benefit can distort prescribing decisions, coverage determinations, or a patient's understanding of their own treatment.

Despite this risk, no standardized framework exists for systematically evaluating how faithfully LLMs summarize clinical trial results across multiple stakeholder audiences. Existing NLI-based faithfulness metrics, developed for general-purpose summarization~\cite{jia2023zeroshot, zhang2024finegrained}, have not been validated against the failure modes most consequential to healthcare: unsupported claims, selective reporting, and statistical misrepresentation.

This paper makes three contributions. First, we introduce a benchmark evaluation framework consisting of 200 stratified trials drawn from the Aggregate Analysis of ClinicalTrials.gov (AACT) database, evaluated across three stakeholder audiences using \linebreak audience-specific prompt templates and a six-dimension faithfulness annotation schema scored on a one-to-three ordinal scale. Second, we apply this framework to establish baseline faithfulness measurements across three leading LLMs, GPT-4o, Claude Sonnet 4.6, and Gemini 2.5 Flash, generating 1,800 summaries scored by a cross-encoder NLI model (DeBERTa-v3-small)~\cite{he2021deberta}, finding that Unsupported Claims is the dominant failure mode across all three models, with a mean annotation score of 1.55 out of three. Third, we introduce a Knowledge Graph Retrieval-Augmented Generation (KG-RAG) system grounded in the PubMed Knowledge Graph (PKG) developed by Xu et al.~\cite{xu2020pubmed}, using hybrid retrieval combining PubMedBERT dense similarity search with PKG concept grounding to augment LLM prompts with relevant biomedical literature. KG-RAG produces statistically significant improvements in overall NLI faithfulness scores (entailment $+0.0125$, faithfulness $+0.0130$, $p < 0.0001$), with consistent directional improvement across all three models.

The remainder of this paper is organized as follows. Section~2 reviews related work across clinical trial summarization, LLM faithfulness evaluation, and KG-RAG in biomedical NLP\@. Section~3 describes the methodology. Section~4 presents results. Sections~5 and~6 provide the discussion and conclusion.

\section{Related Work}

Summarization of clinical trial content has been explored in the NLP literature, although prior work has focused primarily on structured information extraction and extractive summarization rather than a faithfulness evaluation of the generated outputs. Much of the early work demonstrated that automated methods could extract and summarize structured information from clinical trial text, establishing a foundation for clinical trial text processing~\cite{gulden2019extractive,alhussaini2022ccs}. More recent work has applied LLMs directly to clinical regulatory documents, including generating safety summaries for clinical study reports and measuring factual accuracy against the source tables for a single regulatory audience~\cite{landman2024llm}. While these studies demonstrated the feasibility of automated trial summarization, they did not evaluate how faithful the summarization was across multiple stakeholder audiences such as patients, clinicians, and payers, as each requires different framings of the same trial results. Additionally, they did not provide a reusable evaluation framework applicable across all models and therapeutic areas. This study addresses that gap.

NLI has become the preferred automated proxy for the faithfulness evaluation of AI-generated summarization~\cite{zhang2024finegrained}. The core approach treats the source document from the clinical trials as a premise and each generated sentence as a hypothesis, using entailment probability as a measure of factual grounding. Prior work has demonstrated that NLI-based zero-shot faithfulness metrics correlate meaningfully with human judgments across abstractive summarization tasks, offering a scalable alternative to annotation-intensive evaluation pipelines~\cite{jia2023zeroshot,zhang2024finegrained}. This study applies a cross-encoder NLI model (DeBERTa-v3-small) at the sentence level across 1,800 LLM-generated summaries, using \texttt{entailment\_mean} as the primary faithfulness metric.

KG-RAG has shown performance improvements across biomedical NLP tasks~\cite{soman2023biomedical,mortezaagha2026graph}. Xu et al.\ introduced the PKG, which is a large-scale resource that links PubMed abstracts through BioBERT-extracted biomedical entities, including drugs, diseases, and genes, and provides structured concept-level grounding over the biomedical literature~\cite{xu2020pubmed}. Further research has demonstrated that the integration of knowledge graphs into RAG pipelines produces more factually grounded outputs than dense retrieval alone~\cite{soman2023biomedical,mortezaagha2026graph}. This study incorporates PKG2020S4 as the concept-grounding layer of a hybrid retrieval system, pairing PKG entity matching with dense PubMedBERT embeddings to retrieve evidence-rich context for clinical trial summaries. The result is a statistically significant improvement across all three models, with overall entailment increasing by 0.0125 and faithfulness score increasing by 0.0130 ($p < 0.0001$ for both).

\section{Methodology}

\subsection{Dataset}

This study utilizes two publicly available data sources, the AACT database and PubMed abstracts retrieved via the NCBI Entrez API\@. AACT provides the structured trial records that constitute the ground truth for faithfulness evaluation; the PubMed abstracts form the retrieval corpus used by the KG-RAG system.

AACT, which is maintained by the Clinical Trials Transformation Initiative, provides structured flat-file exports of the full ClinicalTrials.gov registry~\cite{tasneem2012aact}. The faithfulness evaluation requires that trial data include actual outcome data; therefore, only studies that were labeled as \textsc{completed} and had a posted results date were used. Each record was then tagged by therapeutic area using condition keywords from the AACT conditions table. Three therapeutic areas that represent meaningfully different clinical communication challenges were selected: oncology, mental health, and type~2 diabetes. Oncology trials tend to involve complex multi-arm designs and molecularly targeted therapies; mental health trials rely heavily on patient-reported outcome measures; and type~2 diabetes trials typically follow well-standardized endpoint conventions. After filtering and tagging, 15,546 trial records remained across the three areas: 11,400 oncology, 2,634 mental health, and 1,512 type~2 diabetes.

Each trial record contributed seven structured fields to the prompt templates used throughout the evaluation: brief title, brief summary, study phase, enrolled participant count, intervention names, primary outcomes, and outcome results. These fields collectively represent the information available to a medical writer or AI system when generating a stakeholder-facing summary. The inclusion of outcome results distinguishes this dataset from protocol-only registries and makes it possible to evaluate whether a generated summary accurately represents trial results as opposed to what the trial intended to measure.

PubMed abstracts were retrieved via the NCBI Entrez API to serve as the retrieval corpus for the KG-RAG system. Rather than fetching the full PubMed archive, a targeted approach was used in which AACT's study references table identified the PubMed record identifiers (PMIDs) corresponding to published results papers for each trial in scope, and only those abstracts were fetched~\cite{ncbi2010entrez}. This produced 13,746 unique abstracts linked to 7,510 trials in the dataset, each record containing a title, abstract text, publication year, journal name, and MeSH terms. These abstracts represent the published evidence base that is most directly relevant to the trials that were to be evaluated and serve as the pool from which the hybrid retriever selects supporting context documents.

\begin{figure}
  \centering
  \includegraphics[width=\linewidth]{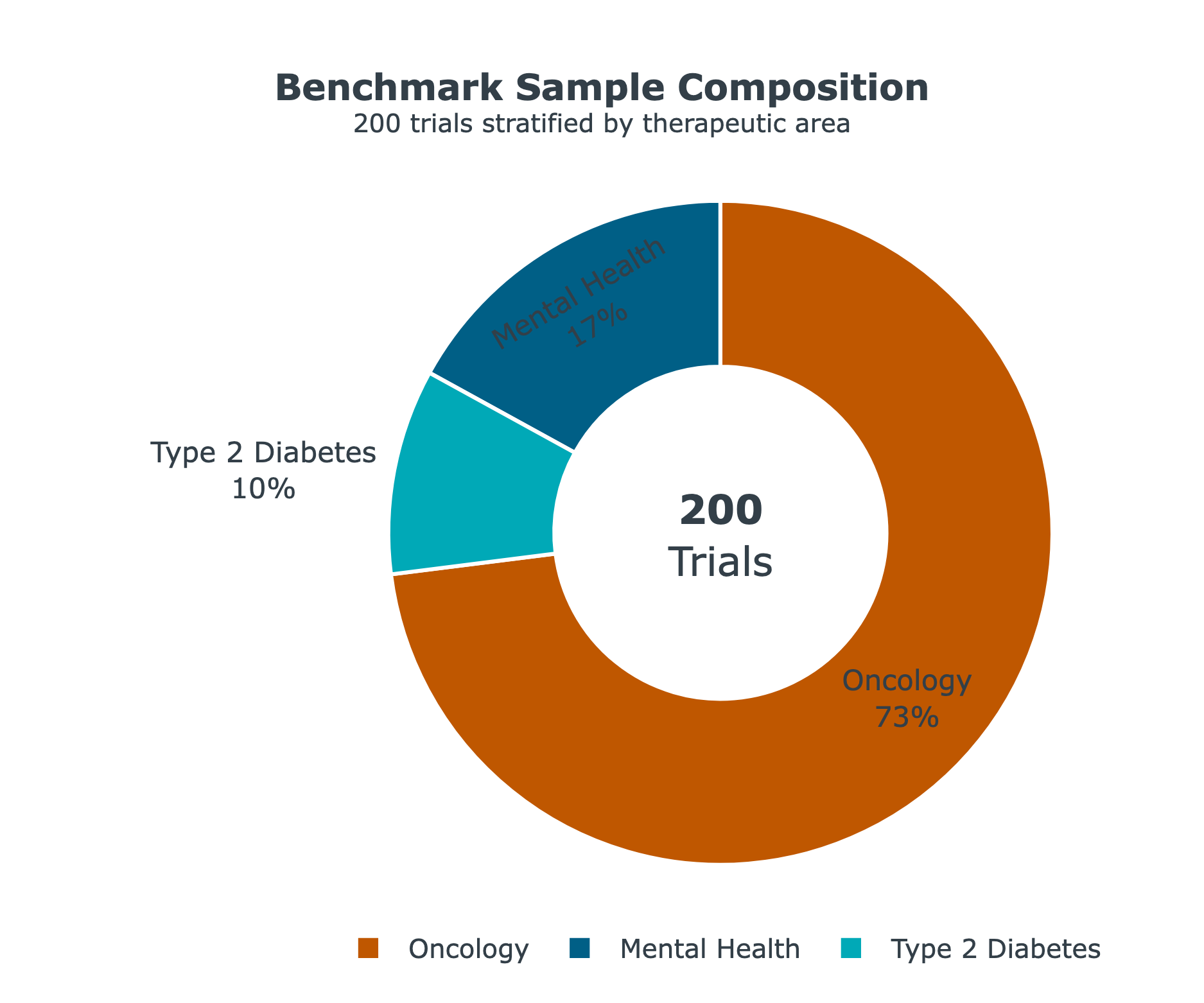}
  \caption{Benchmark composition by therapeutic area ($n = 200$ trials): 146 oncology (73\%), 34 mental health (17\%), and 20 type~2 diabetes (10\%).}
  \label{fig:benchmark}
\end{figure}

To make the evaluation computationally feasible, we constructed a stratified benchmark of 200 trials from the 15,546-record pool. Stratified random sampling was applied across therapeutic areas to preserve the registry's natural distribution: 146 oncology trials, 34 mental health trials, and 20 type~2 diabetes trials. Before sampling, a quality filter was applied requiring each selected trial to have a non-null summary, primary outcomes field, and outcome results field, ensuring that every prompt in the evaluation was complete and that no summary was generated from a partial record. All sampling used a fixed random seed of 42 to ensure reproducibility across runs. Figure~\ref{fig:benchmark} presents the benchmark composition by therapeutic area. Code implementing the full evaluation pipeline is available at \url{https://github.com/rgw3/ai_in_healthcare_hrp}.

\subsection{Evaluation Framework}

The evaluation framework is organized around three stakeholder audiences, each of which represents a distinct communication context for clinical trial results. Healthcare providers require summaries that support clinical decision-making; for this audience, faithfulness focuses on the accurate representation of efficacy outcomes, adverse events, dosing context, and patient selection criteria. Patients require plain-language summaries that support informed consent~\cite{donner2025patient}; therefore, the faithfulness for this audience centers on the absence of unsupported claims, the comparison of risks and benefits, and the avoidance of language that minimizes or obscures adverse findings. Payers require summaries that support coverage and formulary decisions, meaning that faithfulness for this audience centers on the accurate representation of primary endpoints, statistical significance, and the comparative effectiveness framing of the trial's results.~\cite{amcp2024format}

To operationalize faithfulness evaluation across these three audiences, we developed a six-dimension annotation schema. These dimensions were motivated by two sources: clinical trial reporting standards --- in particular, the CONSORT 2010 guidelines for transparent reporting of outcomes, adverse events, and statistical results~\cite{consort2010} --- and established NLI-based faithfulness evaluation frameworks~\cite{jia2023zeroshot, zhang2024finegrained}. Each dimension is scored on a one-to-three ordinal scale, where a score of three indicates full faithfulness, a score of two indicates a minor issue such as a small omission or imprecision that does not materially mislead, and a score of one indicates a clear faithfulness violation such as a fabricated result, an unsupported claim, or a material omission. The six dimensions are: \textit{Factual Accuracy}, defined as the correct representation of numerical results including effect sizes, percentages, and $p$-values; \textit{Outcome Coverage}, defined as the presence of primary outcomes and the absence of key omissions; \textit{Risk and Safety Representation}, defined as the accurate and proportional reporting of adverse events and safety signals; \textit{Statistical Integrity}, defined as the correct conveyance of statistical claims including significance and confidence intervals; \textit{Unsupported Claims}, defined as the absence of claims that go beyond what the trial data supports; and \textit{Selective Reporting}, defined as the balanced representation of positive and negative findings without selectively emphasizing favorable results.

We constructed audience-specific prompt templates for each of the three audiences. Each template instructs the model to generate a summary of 150 to 200 words using only the information present in the provided trial record. The template specifies the audience's decision-making context, explicitly prohibits the introduction of information not contained in the source record, and instructs the model to acknowledge missing data rather than infer or fabricate results. A structured trial context block formatted from seven AACT fields (brief title, brief summary, study phase, enrolled participant count, intervention names, primary outcomes, and outcome results) is provided as the sole source document. All LLM calls were made at a temperature of zero to ensure deterministic and reproducible outputs across runs.

\subsection{Baseline LLM Evaluation}

The baseline evaluation applied each of the three audience-specific prompt templates to all 200 benchmark trials using three large language models: GPT-4o (OpenAI), Claude Sonnet 4.6 (Anthropic), and Gemini 2.5 Flash (Google). This produced 1,800 LLM-generated summaries in total (200 trials $\times$ three audiences $\times$ three models), each generated independently and stored alongside its source trial record, audience label, and model identifier.

We scored all 1,800 summaries for faithfulness using the \texttt{cross-\allowbreak encoder/ nli-deberta-v3-small} model~\cite{he2021deberta}, a cross-encoder NLI model that scores sentence-level textual entailment between a premise and a hypothesis. For each summary, the trial record was formatted as the NLI premise and each sentence of the generated summary was treated as a hypothesis. Sentences shorter than ten characters were excluded to remove fragments. We computed three aggregate metrics per summary: \texttt{entailment\_mean}, defined as the mean entailment probability across all scored sentences; \texttt{contradiction\_mean}, defined as the mean contradiction probability across all scored sentences; and \texttt{faithfulness\_score}, defined as the fraction of sentences for which the entailment probability exceeded 0.5. We ran NLI inference on-device using Metal Performance Shaders acceleration and resolved label indices at runtime from the model's \texttt{id2label} mapping to ensure that the entailment and contradiction class indices were correctly identified regardless of checkpoint ordering.

\subsection{KG-RAG System}

Prior work has demonstrated that retrieval-augmented generation can improve factual grounding in biomedical NLP tasks by providing language models with grounding context from relevant source documents~\cite{soman2023biomedical,mortezaagha2026graph}. For clinical trial summarization, the natural grounding source is the peer-reviewed literature directly linked to each trial. This study builds on that approach, using the PKG developed by Xu et al.~\cite{xu2020pubmed} as the concept-grounding layer of a hybrid retrieval architecture designed to augment LLM prompts with evidence from the trial-linked abstract corpus.

PKG2020S4, available at \url{https://er.tacc.utexas.edu/datasets/ped}, covers PubMed publications through December 2020 and provides BioBERT-extracted biological entity annotations per PMID, including drugs, diseases, genes, species, and mutations. This study uses the drug and disease entity types from the PKG entity file, which contains 122,884 filtered entity mentions across 7,812 of the 13,746 trial-linked abstracts (56.8\%). The remaining 43.2\% of the abstract corpus falls outside PKG2020S4's December 2020 coverage window, a hard recall ceiling that reflects the static nature of the knowledge graph snapshot and is acknowledged as a limitation of the system. An initial design based on pure concept-lookup retrieval, in which drug and disease mentions from each trial record were matched against the PKG entity index to identify candidate abstracts, revealed severe coverage gaps in practice: trastuzumab returned two matching PMIDs, while pembrolizumab and bevacizumab returned zero. These gaps stem from both the temporal cutoff and the limited coverage of brand and generic drug-name variants in BioBERT extraction. This finding motivated a shift to hybrid retrieval, in which dense embedding similarity serves as the primary retrieval backbone and PKG concept matching provides an additive re-ranking signal.

The dense retrieval component encodes all 13,746 trial-linked abstracts using the \texttt{pritamdeka/S-PubMedBert-MS-MARCO} sentence encoder, a PubMedBERT-based model fine-tuned for biomedical retrieval~\cite{gu2021pubmedbert}. Each abstract is represented as the concatenation of its title and abstract text and projected to a 768-dimensional L2-normalized vector. Retrieval proceeds in three stages for each benchmark trial: the trial's brief title and brief summary are encoded as a query vector, cosine similarity is computed against the full abstract matrix to produce a candidate pool of 200 abstracts, and a PKG concept-grounding boost of 0.15 is added to the final score of any candidate whose PMID is linked to a matched surface form from the trial record. The top ten abstracts by boosted score are selected as the retrieval result. This retrieve-once-per-trial design ensures that the same ten abstracts are provided as context for all nine audience-and-model combinations derived from a given trial, guaranteeing identical evidence inputs for a fair cross-model comparison.

Retrieved abstracts are injected into the prompt as a numbered evidence block placed before the trial record. Each entry includes the abstract's rank, PMID, and title as a header followed by up to 1,500 characters of abstract text. A grounding instruction prepended to the prompt directs the model to use retrieved abstracts only as supporting background context and explicitly prohibits introducing information not present in the trial record. This preserves the trial record as the sole authoritative source while providing the additional biomedical context that may help anchor the model's claims.

\begin{figure}
  \centering
  \includegraphics[width=\linewidth]{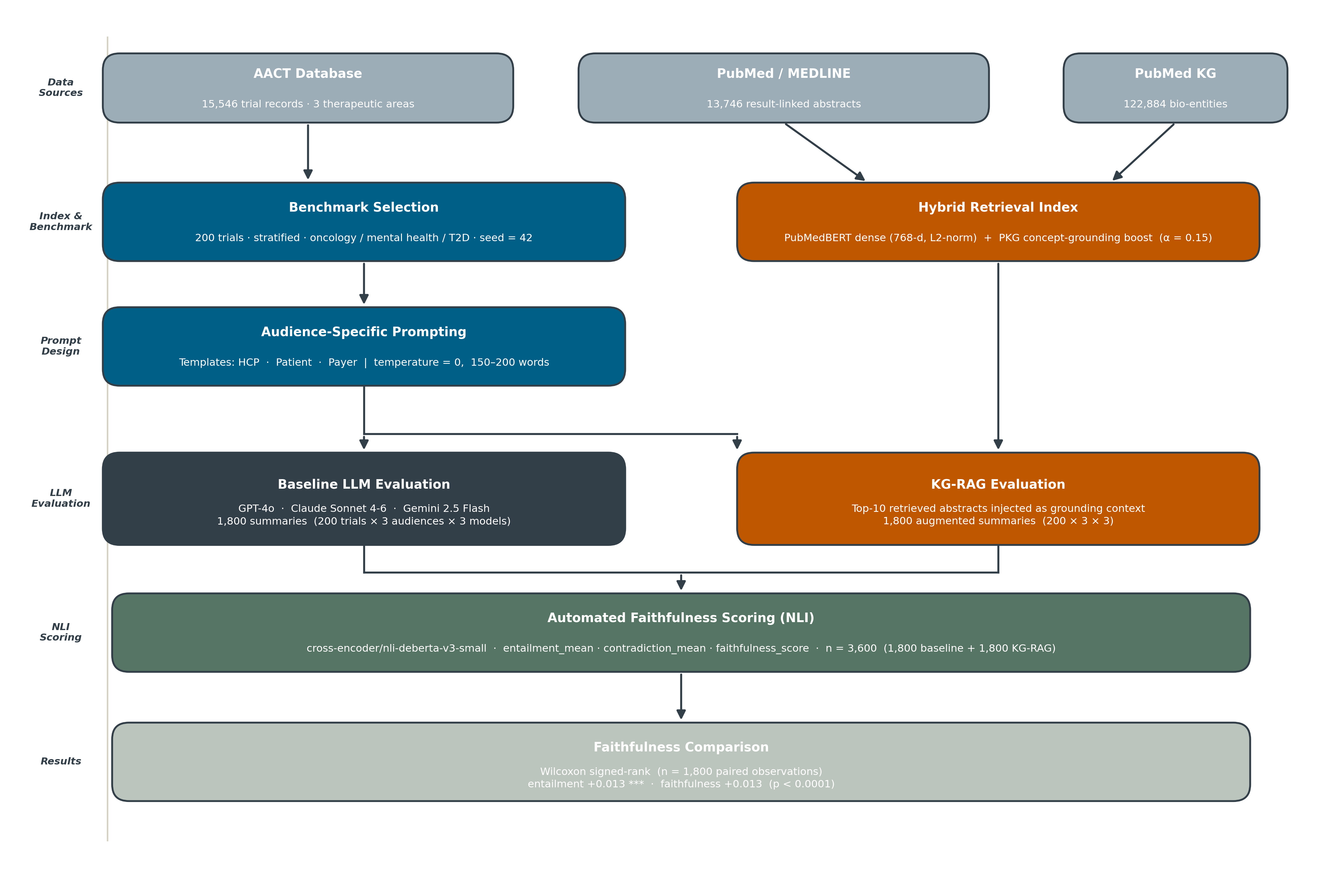}
  \caption{Overview of the complete evaluation pipeline, from AACT and PubMed data sources through KG-RAG augmentation and NLI scoring.}
  \label{fig:pipeline}
\end{figure}

The KG-RAG evaluation applied the same 200 benchmark trials, three audiences, and three models as the baseline, producing 1,800 augmented LLM-generated summaries scored using the same NLI pipeline described in Section~3.3. The trial record served as the NLI premise in both conditions so that the comparison measures faithfulness to the source record rather than alignment with the retrieved abstracts. Statistical comparison between baseline and KG-RAG NLI scores used the Wilcoxon signed-rank test on $n = 1{,}800$ paired observations per metric, a non-parametric test appropriate for paired continuous measurements where normality cannot be assumed. Figure~\ref{fig:pipeline} presents a visual overview of the complete evaluation pipeline, from data sources through NLI scoring.

\section{Results}

\subsection{Baseline Annotation Results}

\begin{figure}
  \centering
  \includegraphics[width=\linewidth]{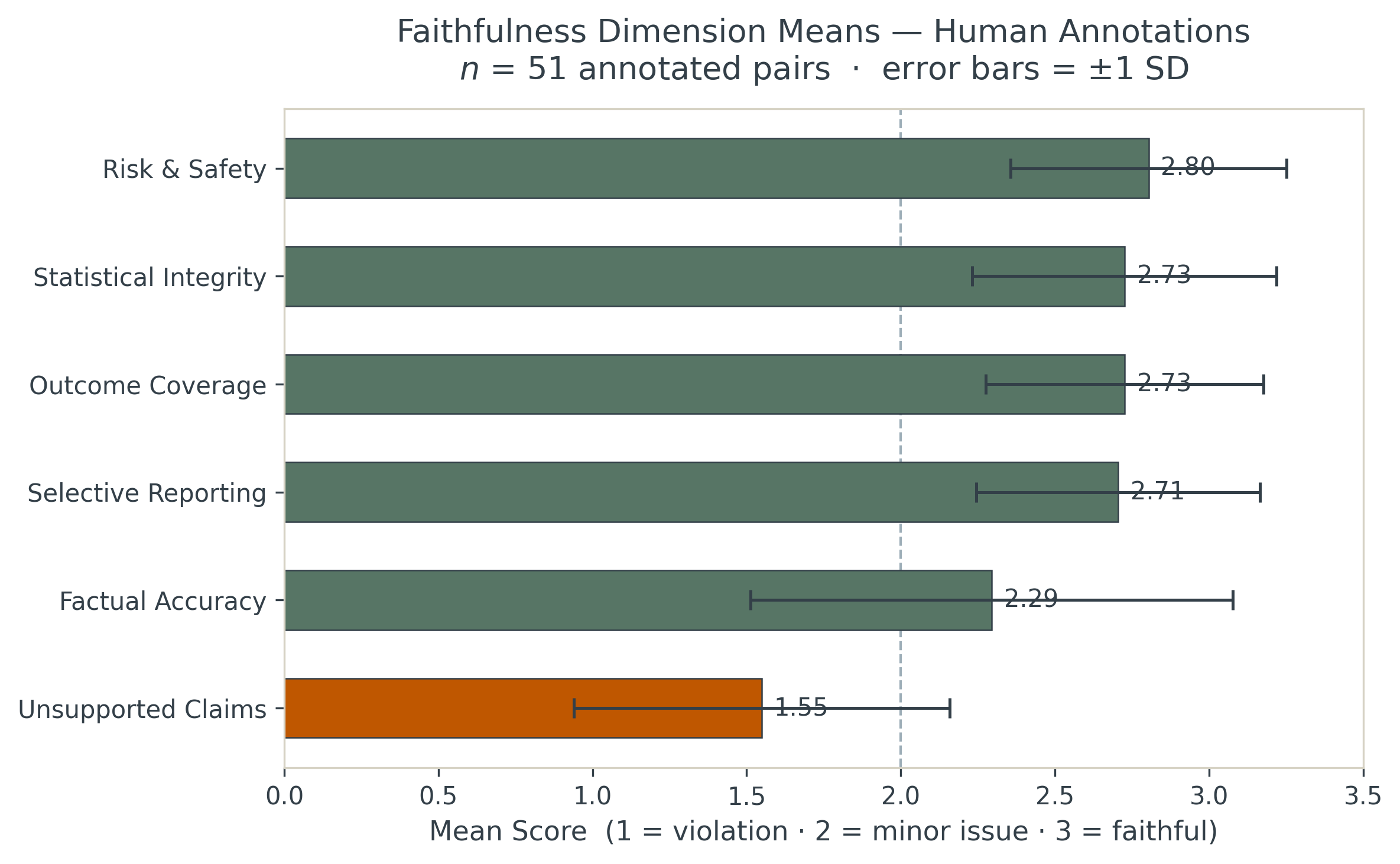}
  \caption{Mean annotation scores and standard deviations across the six faithfulness dimensions ($n = 51$ annotated pairs). Unsupported Claims (1.55) was the dominant failure mode.}
  \label{fig:dimensions}
\end{figure}

A single annotator scored 51 baseline output pairs across all six faithfulness dimensions using the annotation framework described in Section~3.2. Scores are reported on a one-to-three scale, where three indicates full faithfulness and one indicates a clear faithfulness violation. Figure~\ref{fig:dimensions} presents the mean score and standard deviation for each dimension across all 51 pairs.

Across all six dimensions, Unsupported Claims produced the lowest mean score (1.55), identifying it as the dominant failure mode in baseline LLM-generated clinical trial summaries. This finding indicates that all three models regularly introduced claims that could not be traced to the source trial record, regardless of audience. Factual Accuracy produced the second-lowest mean (2.29), indicating that numerical results in generated summaries were frequently imprecise or incorrect. Selective Reporting scored 2.71, suggesting that models tended to present trial results in a directionally favorable light. Statistical Integrity scored comparatively higher (2.73), indicating that when statistical claims were present in the source, models generally conveyed them correctly, though imprecision remained common.

\begin{figure}
  \centering
  \includegraphics[width=\linewidth]{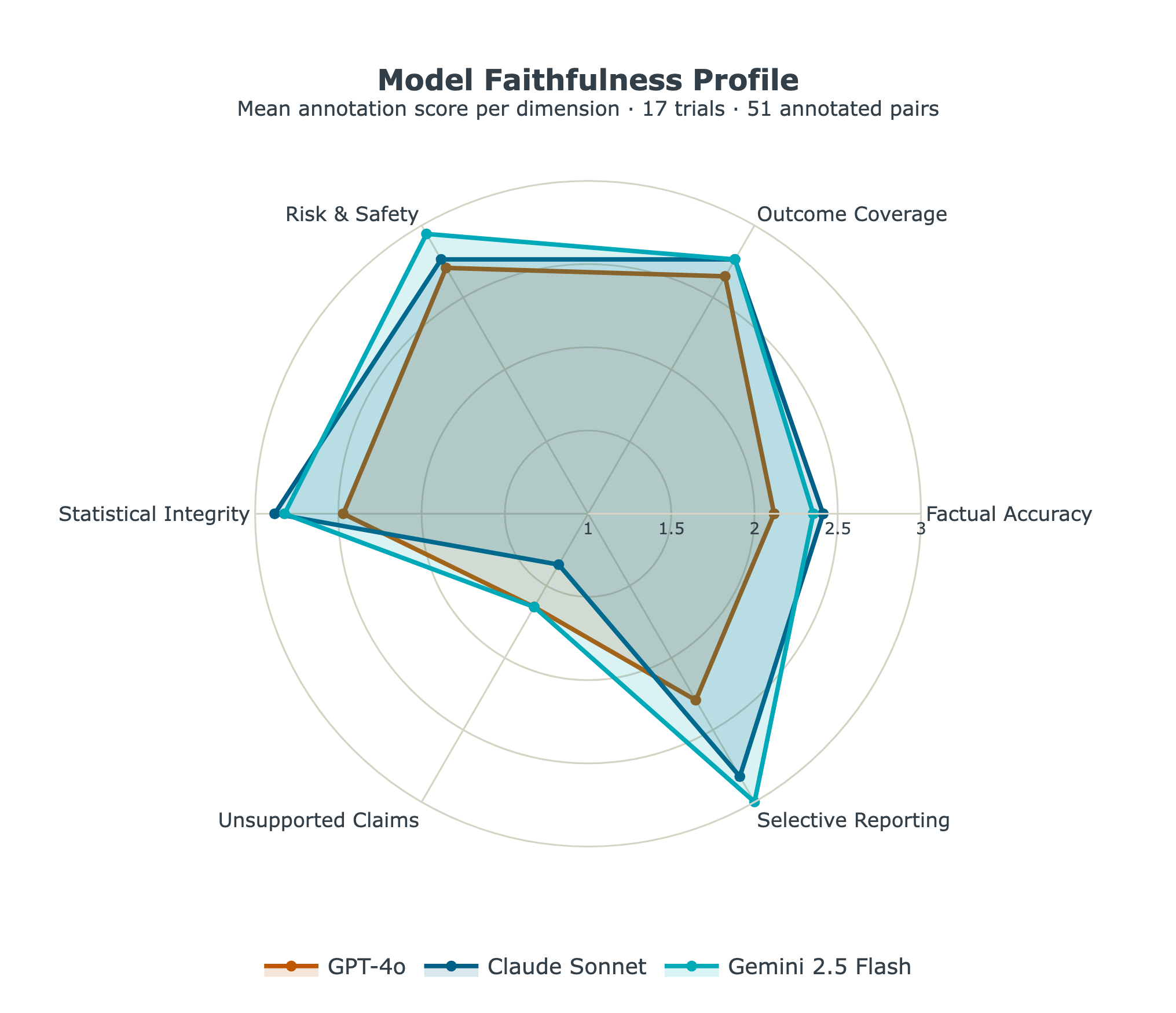}
  \caption{Radar chart of mean annotation scores by model across the six faithfulness dimensions. GPT-4o, Claude Sonnet 4.6, and Gemini 2.5 Flash show broadly similar profiles.}
  \label{fig:radar}
\end{figure}

\begin{figure}
  \centering
  \includegraphics[width=\linewidth]{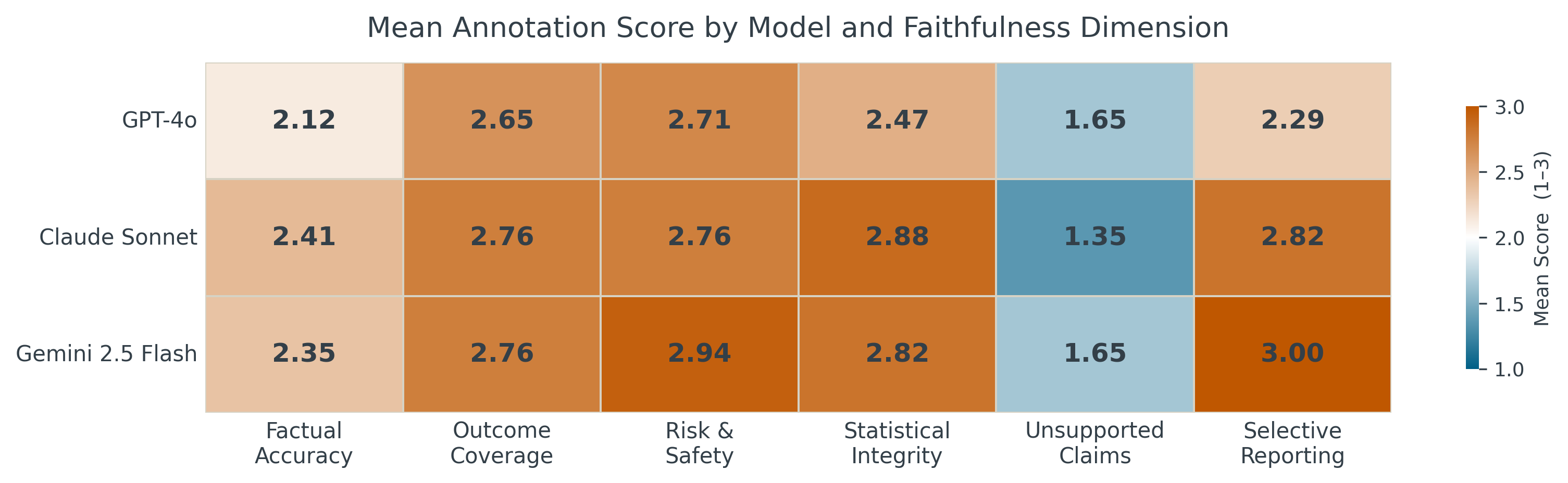}
  \caption{Heatmap of mean faithfulness scores disaggregated by model and audience across the six annotation dimensions.}
  \label{fig:heatmap}
\end{figure}

Figure~\ref{fig:radar} presents a radar chart overlaying all three models across the six dimensions. GPT-4o, Claude Sonnet 4.6, and Gemini 2.5 Flash produced broadly similar profiles, with the largest divergence appearing on the Unsupported Claims and Selective Reporting dimensions. Figure~\ref{fig:heatmap} presents a heatmap of mean scores disaggregated by model and audience. Scores were generally consistent across the three audience types, with no single audience systematically producing more faithful summaries than another. The patient audience showed a slight tendency toward lower Unsupported Claims scores, consistent with the expectation that plain-language summaries require more paraphrasing and inference from the model.

\subsection{NLI Metric Validation}

\begin{figure}
  \centering
  \includegraphics[width=\linewidth]{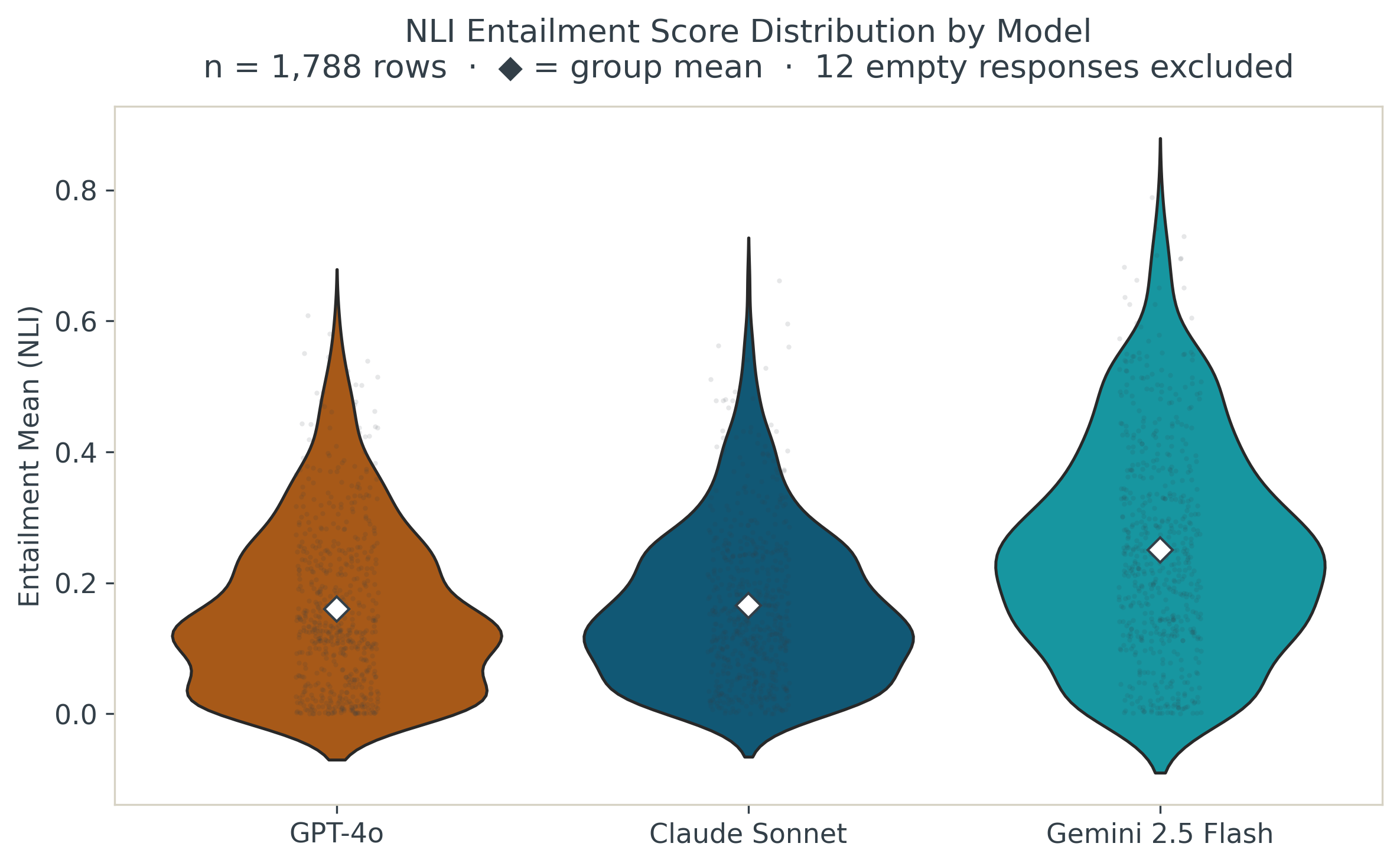}
  \caption{Violin plots of \texttt{entailment\_mean} score distributions for each model across 600 summaries per model. All three models show means well below 0.5 at baseline.}
  \label{fig:violin}
\end{figure}

Figure~\ref{fig:violin} presents violin plots of the \texttt{entailment\_mean} NLI score distributions for each of the three models across 600 summaries per model. The distributions overlap substantially, indicating that baseline faithfulness levels are comparable across GPT-4o, Claude Sonnet 4.6, and Gemini 2.5 Flash at the automated metric level. All three distributions show \texttt{entailment\_mean} values well below 0.5, with means of 0.160 (GPT-4o), 0.165 (Claude Sonnet 4.6), and 0.251 (Gemini 2.5 Flash), indicating that most generated sentences are only weakly grounded in the source record at baseline.

\begin{figure}
  \centering
  \includegraphics[width=\linewidth]{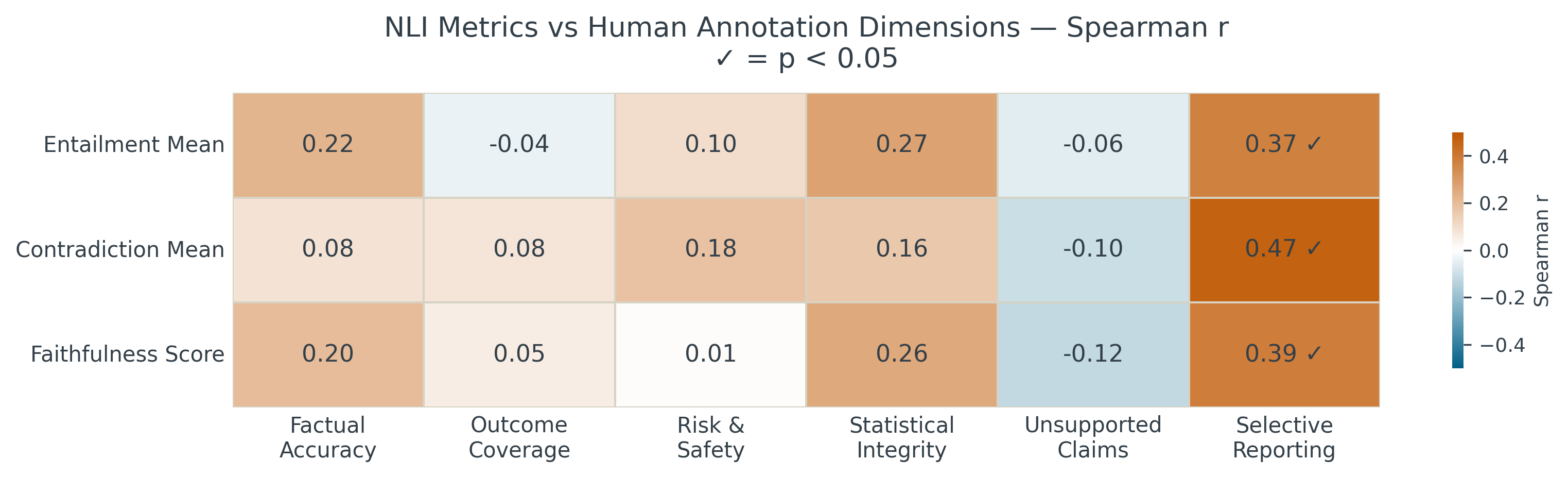}
  \caption{Spearman rank correlation matrix between NLI metrics and human annotation dimensions across 51 annotated pairs. \texttt{entailment\_mean} shows the strongest composite correlation ($r = +0.304$, $p = 0.030$).}
  \label{fig:correlation}
\end{figure}

To validate the NLI metrics against the human annotation results, we computed Spearman rank correlations between each of the three NLI metrics and each of the six human-scored dimensions across the 51 annotated pairs. Figure~\ref{fig:correlation} presents the resulting correlation matrix. Of the three NLI metrics, \texttt{entailment\_mean} produced the strongest composite correlation with human faithfulness scores ($r = +0.304$, $p = 0.030$), supporting its use as the primary automated metric throughout this study. \texttt{Contradiction\_mean} and \texttt{faithfulness\_score} produced weaker and less consistent correlations across dimensions.

The correlation between NLI metrics and the Unsupported Claims dimension was the weakest observed in the matrix. This finding is consistent with the nature of unsupported claims as a failure mode: NLI entailment measures whether a sentence is supported by the premise, but it does not reliably detect claims that are plausible, coherent, and factually unrelated to the source. A summary sentence that introduces a mechanism of action not present in the trial record may receive a high entailment score because it does not contradict the premise, even though the claim cannot be traced to the source record. This gap between what NLI measures and what the Unsupported Claims dimension captures is acknowledged as a limitation of the automated evaluation approach.

\subsection{KG-RAG vs.\ Baseline}

\begin{figure}
  \centering
  \includegraphics[width=\linewidth]{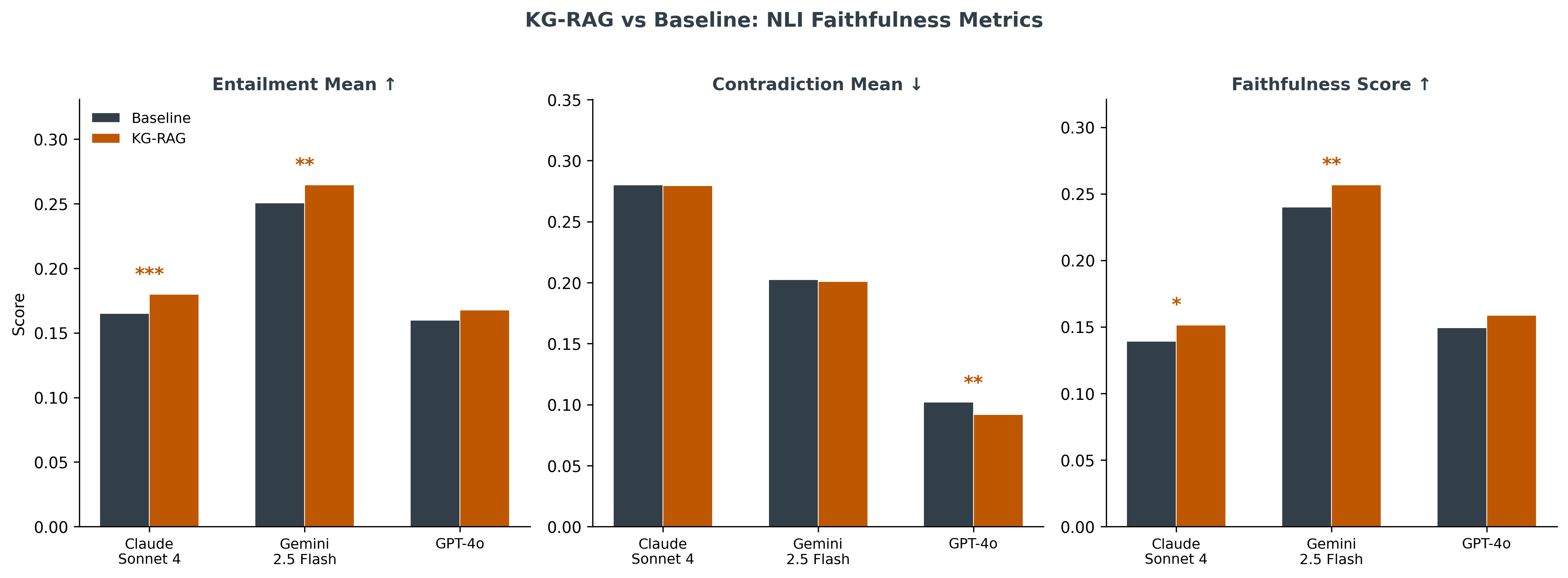}
  \caption{Mean NLI scores for baseline and KG-RAG conditions by model. KG-RAG produces statistically significant improvements in \texttt{entailment\_mean} ($+0.0125$, $p < 0.0001$) and \texttt{faithfulness\_score} ($+0.0130$, $p < 0.0001$).}
  \label{fig:kgrag}
\end{figure}

The KG-RAG system was evaluated against the baseline across the same 200 benchmark trials, three audiences, and three models, producing 1,800 paired NLI scores per metric. Figure~\ref{fig:kgrag} presents, by model, the mean \texttt{entailment\_mean}, \texttt{contradiction\_mean}, and \texttt{faithfulness\_score} for baseline and KG-RAG conditions.

Wilcoxon signed-rank tests on the 1,800 paired observations yielded statistically significant improvements on two of the three metrics. \texttt{Entailment\_mean} increased by 0.0125 ($p < 0.0001$) and \texttt{faithfulness\_score} increased by 0.0130 ($p < 0.0001$). These results indicate that augmenting LLM prompts with PKG-grounded PubMed abstracts produces measurably more faithful summaries relative to source trial records, at a scale sufficient to detect the effect with high statistical confidence.

The improvement pathways differed across models. Claude Sonnet 4.6 and Gemini 2.5 Flash showed improvement primarily through increased entailment, suggesting that the retrieved context helped these models generate sentences that were more directly grounded in the trial record. GPT-4o showed a different pattern: its primary improvement was a reduction in \texttt{contradiction\_mean} of 0.010 ($p = 0.001$), with a smaller entailment gain. This suggests that for GPT-4o, the retrieved context was more effective at suppressing contradictory outputs than at increasing positive grounding. The practical implication is that KG-RAG improves faithfulness across all three models, but the mechanism by which it does so is model-dependent.

\section{Discussion}

The most consistent finding across the baseline evaluation is that Unsupported Claims was both the dominant annotation failure mode, with a mean score of 1.55, and the dimension most resistant to automated detection by NLI scoring. The Spearman correlation between \texttt{entailment\_mean} and Unsupported Claims scores was effectively zero ($r = -0.058$, $p = 0.686$), and no NLI metric produced a meaningful correlation with this dimension. This result is explained by the nature of the failure mode itself. NLI entailment measures whether a generated sentence is logically consistent with the source record; it does not measure whether a sentence originates from the source record. A hallucinated claim that is biomedically plausible and does not directly contradict the trial data will receive a normal entailment score even though no part of the trial record supports it. This distinction between non-contradiction and positive grounding is consequential for clinical summarization, where the absence of a finding in the source is itself meaningful information. The practical implication is that NLI-based evaluation, while useful as a scalable faithfulness proxy, cannot fully substitute for human review in detecting unsupported claims.

The 95\% confidence intervals on Spearman $r$ (estimated via Fisher's $z$-transformation) reflect the constraints of the 51-pair annotation sample. For \texttt{entailment\_mean}, the composite interval spans $[{+0.03}, {+0.53}]$, confirming a positive directional relationship while acknowledging meaningful uncertainty in effect magnitude. The \texttt{contradiction\_mean} composite interval ($[-0.05, +0.48]$) is consistent with the non-significant $p$-value ($p = 0.105$) and should be interpreted as directional evidence only. By contrast, \texttt{contradiction\_mean}'s correlation with Selective Reporting, the strongest individual finding, produces a narrower interval ($[{+0.22}, \allowbreak {+0.66}]$) that is entirely positive, supporting the claim that NLI contradiction signals are particularly sensitive to one-sided reporting. The annotation sample is sufficient to establish the directional validity of \texttt{entailment\_mean} as a composite faithfulness proxy, but magnitude estimates should be treated as preliminary pending a larger annotation study. NLI metrics are best understood as a screening tool for directional faithfulness trends rather than a calibrated measurement instrument, particularly given their inability to detect Unsupported Claims, the most clinically consequential failure mode.

The KG-RAG results indicate that retrieval augmentation improves faithfulness across all three models, but the mechanism differs. Claude Sonnet 4.6 and Gemini 2.5 Flash improved primarily through increased entailment, with deltas of $+0.0147$ and $+0.0141$, respectively, and negligible changes in contradiction ($-0.0011$ and $-0.0029$). GPT-4o showed the opposite pattern: its entailment gain was smaller ($+0.0079$) while its contradiction reduction was the largest of the three models ($-0.0100$, $p = 0.001$). This suggests that Claude Sonnet 4.6 and Gemini 2.5 Flash used the retrieved context to generate sentences more directly grounded in the biomedical literature, while GPT-4o used it primarily to suppress outputs that conflicted with the retrieved evidence. Whether this reflects a difference in how each model weights retrieved context relative to parametric knowledge is a question that the current study design cannot resolve, but the pattern is consistent across 600 summaries per model and warrants further investigation.

The PKG2020S4 coverage ceiling represents a structural limitation of the current system. Of the 13,746 trial-linked abstracts, only 56.8\% contained PKG entity annotations, and the remaining 43.2\% fell outside the knowledge graph's December 2020 temporal cutoff. For recent trials involving therapies approved or studied after that date, the concept-grounding boost provided no signal, and retrieval fell back entirely on dense embedding similarity. Given that the benchmark includes oncology trials, a therapeutic area characterized by a high rate of novel targeted therapies, a more current knowledge graph snapshot would likely improve both coverage and retrieval precision for this subset of trials.

Several limitations of this study should be noted. The NLI validation relied on 51 annotation pairs scored by a single annotator, which limits the statistical power of the correlation analysis and the generalizability of the validation finding. The study evaluated three therapeutic areas selected to represent meaningfully different clinical communication challenges; performance on other therapeutic areas was not assessed. Finally, as discussed above, the NLI evaluation framework cannot reliably detect the Unsupported Claims failure mode that this study identifies as the most prevalent. Future work incorporating a larger, multi-annotator evaluation with calibrated inter-annotator agreement would provide stronger validation of the automated metrics and a more precise estimate of the KG-RAG system's effect on this specific dimension.

\section{Conclusion}

This study introduced a benchmark evaluation framework for measuring the faithfulness of LLM-generated clinical trial summaries across multiple stakeholder audiences. The framework consists of 200 stratified trials drawn from the AACT database, evaluated across three audience personas using audience-specific prompt templates and a six-dimension annotation schema scored on a one-to-three ordinal scale. Baseline faithfulness measurements were established for GPT-4o, Claude Sonnet 4.6, and Gemini 2.5 Flash across 1,800 generated summaries, and a KG-RAG system grounded in the PubMed Knowledge Graph was developed and evaluated against that baseline.

The central finding of this study is that KG-RAG augmentation produces statistically significant improvements in overall NLI-based faithfulness scores (\texttt{entailment\_mean} $+0.0125$, \texttt{faithfulness\_\allowbreak score} $+0.0130$, $p < 0.0001$), with consistent directional improvement across all three models. The improvement is consistent in direction across models but model-dependent in mechanism, with Claude Sonnet 4.6 and Gemini 2.5 Flash improving primarily through increased entailment and GPT-4o improving primarily through reduced contradiction. These results support the use of knowledge-graph-augmented retrieval as a practical intervention for improving the faithfulness of LLM-generated clinical summaries in a controlled benchmark setting.

The study also identified a meaningful gap in the automated evaluation framework. Unsupported Claims was the dominant failure mode across all three models, with a mean annotation score of 1.55 out of three, yet no NLI metric produced a meaningful correlation with this dimension ($r = -0.058$, $p = 0.686$ for \texttt{entailment\_mean}). This finding suggests that the most clinically consequential failure mode in LLM-generated trial summaries is also the one least detectable by current automated metrics, and it points to a clear direction for future work.

Several extensions of this study are warranted. First, conducting a multi-annotator human evaluation study with calibrated annotators and sufficient sample size would allow direct testing of whether the NLI-detected KG-RAG improvement is also detectable by trained human evaluators, providing stronger validation of the automated metric. Second, replacing PKG2020S4 with a more current knowledge graph snapshot would remove the December 2020 coverage ceiling that limited entity-grounded retrieval for 43.2\% of the abstract corpus. Third, extending the framework to additional therapeutic areas beyond oncology, mental health, and type~2 diabetes would provide a more complete picture of how LLM faithfulness varies across clinical domains. Collectively, these directions build on the evaluation infrastructure introduced here and move toward a more complete understanding of where and how LLMs fail in high-stakes clinical communication.

\begin{acks}
The author thanks Professor Ying Ding and the UT Austin iSchool for making the PubMed Knowledge Graph (PKG2020S4) publicly available through the Texas Advanced Computing Center.
\end{acks}

 \end{document}